\documentclass[runningheads]{llncs}
\usepackage{graphicx}

\usepackage{tikz}
\usepackage{comment}
\usepackage{amsmath,amssymb} 
\usepackage{color}

\usepackage{algorithm,algpseudocode}
\usepackage{cite}
\usepackage{array}
\usepackage{color}        
\usepackage{mdwmath}
\usepackage{mdwtab}
\usepackage{tabularx, booktabs}
\usepackage{bigstrut}
\usepackage{bm}
\usepackage{epsfig}
\usepackage{subfigure}
\usepackage{wrapfig}
\usepackage{microtype}
\usepackage{multirow}
\usepackage{enumitem}
\usepackage{xspace}
\usepackage{array}
\usepackage{footnote}

\usepackage{expl3}
\ExplSyntaxOn
\newcommand\latinabbrev[1]{
  \peek_meaning:NTF . {
    #1\@}%
  { \peek_catcode:NTF a {
      #1.\@ }%
    {#1.\@}}}
\ExplSyntaxOff

\def\etal{\latinabbrev{.etal}}

\begin{document}
\pagestyle{headings}
\mainmatter
\def\ECCVSubNumber{3242}  

\title{Spherical Feature Transform for Deep Metric Learning} 

\titlerunning{Spherical Feature Transform}
%
\author{Yuke Zhu$^\star$\inst{1}
\and
Yan Bai \thanks{Equal contribution.}
\inst{2}
\and
Yichen Wei\inst{1}
}
\authorrunning{Yuke Zhu et al.}
%
\institute{MEGVII Inc.
\email{\{zhuyuke, weiyichen\}@megvii.com}\\
\and
Tongji Unversity, China
\email{yan.bai@tongji.edu.cn}}
\maketitle

\begin{abstract}
Data augmentation in feature space is effective to increase data diversity. Previous methods assume that different classes have the same covariance in their feature distributions. Thus, feature transform between different classes is performed via translation.  However, this approach is no longer valid for recent deep metric learning scenarios, where feature normalization is widely adopted and all features lie on a hypersphere.

This work proposes a novel spherical feature transform approach. It relaxes the assumption of identical covariance between classes to
an assumption of similar covariances of different classes on a hypersphere. Consequently, the feature transform is performed by a rotation that respects the spherical data distributions. We provide a simple and effective training method, and in depth analysis on the relation between the two different transforms. Comprehensive experiments on various deep metric learning benchmarks and different baselines verify that our method achieves consistent performance improvement and state-of-the-art results.
\end{abstract}

\section{Introduction}

It is crucial to have sufficient data diversity in deep metric learning. A common practice is to augment data in the image space. This is effective but has limited effect. Specifically, it is hard to generate variances in one class using the information in the other classes.

\begin{figure}[!!t]
\begin{center}
\begin{tabular}{c@{}c}
\includegraphics[height=1.8cm]{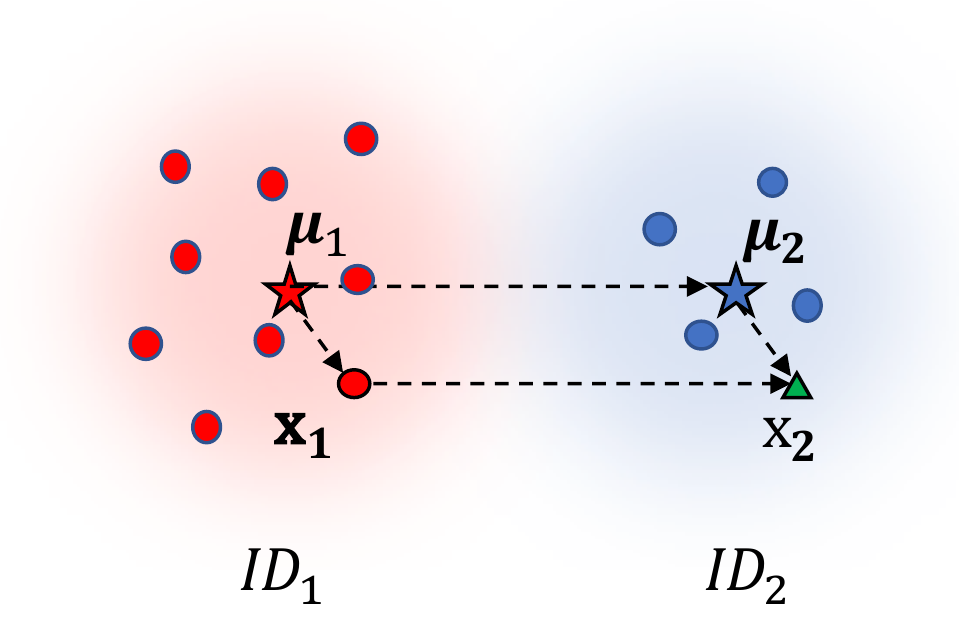} &
\includegraphics[height=1.8cm]{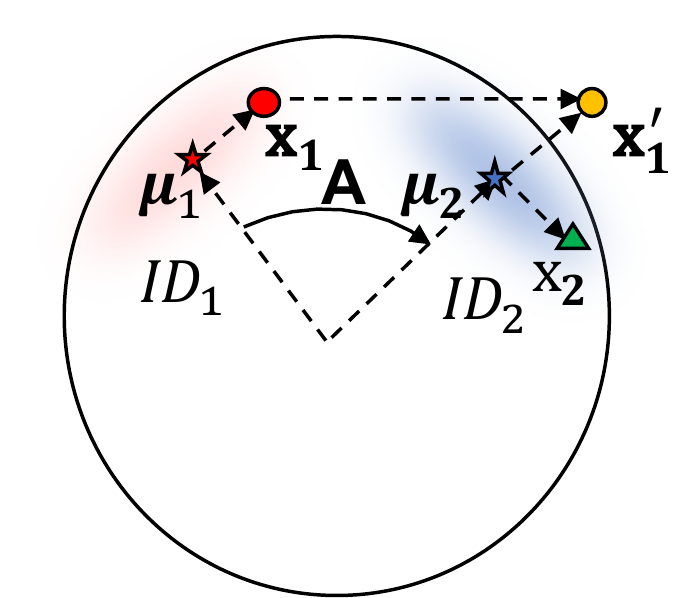}\\
\small (a) & (b)\\ [-1mm]
\end{tabular}
\end{center}
\caption{Illustration of two feature transforms. (a) \textit{translation transform}~\cite{featureTransferLearning}. The feature of $\text{ID}_1$ and $\text{ID}_2$ are sampled from Gaussian distributions with mean value ${\bm \mu}_1$,${\bm \mu}_2$ and identical covariance. To increase the intra-class variances of $\text{ID}_2$, feature ${\bf x}_2$ is generated by translating ${\bf x}_1$ by $\bm{\mu}_2 - \bm{\mu}_1$. (b) Illustration of \emph{translation transform} and SFT on a sphere. Directly translating ${\bf x}_1$ from $\text{ID}_1$ to $\text{ID}_2$ will result in ${\bf x}_1'$, which is out of the surface of the sphere. Our spherical feature transform performs a rotation, such that feature ${\bf x}_1$ of $\text{ID}_1$ is transferred to ${\bf x}_2$ of $\text{ID}_2$.}
\label{fig:illustration}
\end{figure}

Directly augmenting data in the feature space has become a new trend~\cite{duan2018deepAdversarial,zhao2018adversarialApproach,radford2015unsupervised,lin2018deepVariational,featureTransferLearning,liu2018featureSpaceTransfer,zheng2019hardnessHDML}. Specifically, Yin \etal~\cite{featureTransferLearning} propose a simple method that requires no extra labeling and is easy to implement. It assumes that the example features in each class follow a Gaussian distribution, and the covariance between all classes is the same, thus shared. Each feature is the summation of the class-dependent mean and a class-independent variance. Thus, given existing features in one class, their variance parts can be transferred to generate \emph{new} features in other classes, via a translation. This is illustrated in Fig.~\ref{fig:illustration}(a). It is shown effective in~\cite{featureTransferLearning}.

Recently, feature normalization is widely adopted in deep metric learning~\cite{ranjan2017l2L2face,wang2017normface,wang2019ranked,wang2019multi,wang2018cosface,deng2019arcface}. In this case, all features lie on the surface of a hypersphere. The feature transfer approach~\cite{featureTransferLearning} becomes inappropriate. First, a Gaussian distribution is no longer correct. A proper spherical distribution should be used instead. Second, although each class can be approximated as a local Gaussian on the sphere, the assumption of identical covariance between classes is less valid. 
Last, feature translation would produce an invalid feature that is out of the surface of the hypersphere, as shown in Fig.~\ref{fig:illustration}(b). Therefore, both the prior and the feature transform should be adapted for the spherical case.

This work proposes \emph{spherical feature transform} to resolve above problems. It assumes that distributions of features of different classes are spherical-homoscedastic~\cite{hamsici2007spherical}. This relaxes the previous assumption that identical covariance between classes. Instead, it assumes all classes have \emph{similar} covariances, where the similarity is measured by equivalence of eigenvalues of the covariance matrices. Consequently, the transformation between two classes is a rotation that is characterized by the classes' means. This is illustrated in Fig.~\ref{fig:illustration}(b). Theoretical analysis reveals that our approach is a generalization of~\cite{featureTransferLearning}.

Our method is simple and general. It is validated on several deep metric learning tasks. Comprehensive experiments and ablation studies demonstrate its effectiveness.

\section{Related Work}
\label{sec:related}

Feature augmentation is a relatively new topic. 
Some researchers~\cite{duan2018deepAdversarial,zhao2018adversarialApproach, sohn2017unsupervised,zheng2019hardnessHDML} adopt an adversarial approach to generate hard features from the observed negative samples utilizing the Generative Adversarial Networks~(GAN)~\cite{goodfellow2014generativeGAN}. Their main focus is to generate hard negative features. While the structure of feature distributions is not considered. Also, the training process with GAN is usually complicated and unstable~\cite{brock2018largeGANTraining}. Dixit \etal ~\cite{dixit2017aga} propose a data augmentation method using attribute-guided feature descriptor for generation. Liu \etal~\cite{liu2018featureSpaceTransfer} propose to learn a pose manifold in the feature space and use it to synthesize pose-augmented features. However, these works need extra labeling for supervision. 

Recently, Lin \etal ~\cite{lin2018deepVariational} utilize the variational inference to disentangle intra-class variance and leverages the distribution to generate discriminative samples to improve robustness. This work and ours share similar insight that the variances of different class can be regarded as similar. But their method is based on the assumption that the variances can be fully disentangled and can be modeled using a Gaussian. While our method makes no assumptions about this. In fact, we will show that when features are on a hypersphere, the intra-class variances can not be modeled using one distribution. The most similar work to ours is in \cite{featureTransferLearning}. This work also models the variances using a Gaussian. It proposes to transfer the variance part from one class to the other for feature augmentation. It will be detailedly introduced in Sec~\ref{sec:review}. However, both two works do not considering the widely adopted feature normalization techniques and its influence on feature distributions.

\section{Proposed Approach}
\label{sec:approach}
\subsection{Review of Feature Transform}
\label{sec:review}
Feature transform is an approach for feature generation by transferring the intra-class variance from one class to the others. 
It is based on the assumption that features from each class follow a Gaussian distribution and the distributions of different classes have different mean values but shared covariances. Using this assumption, a feature $\bf x$ is represented by two parts:
\begin{equation}
{\bf x} = \bm {\mu + \sigma},
\label{eq:split}
\end{equation}
where $\bm \mu$ is the mean value of the class that ${\bf x}$ belongs to. $\bm \sigma$ is the variance part sampled from a zero-mean Gaussian. $\bm \mu$ contains the information of identity of the class. $\bm \sigma$ contains the information of intra-class variance that is shared among classes.

Following this prior, Feature Transfer Learning~(FTL)~\cite{featureTransferLearning} is proposed to transfer the variance part from one class to the others for feature generation. 
Specifically, given a feature ${\bf x}_1$ with ${\bf x}_1 = \bm {\mu}_1 + \bm{\sigma}_1$ and the center of a target class ${\bm \mu}_2$. The feature generation is proceeded by
${\bf \tilde{x}}_2 = \bm{\mu}_2 + \bm{\sigma}_1$, where ${\bf \tilde{x}}_2$ is regarded as belonging to the target class but shares identical variance with ${\bf x}_1$. We illustrate this process in Fig.~\ref{fig:illustration}(a). The feature transform can also be written as 
\begin{equation}
{\bf \tilde{x}}_2 = {\bf x}_1 + \bm{\mu}_2 - \bm{\mu}_1.
\label{eq:translation}
\end{equation}
It can be interpreted as translating the feature ${\bf x}_1$ by $\bm{\mu}_2 - \bm{\mu}_1$. Thus, this method is referred to as \textit{translation transform}.

\subsection{Review of Spherical-homoscedasticity}
\label{sec:spherical}

Spherical-homoscedasticity is a property describing the relationship between a set of data distributions on the sphere, which we refer to as spherical distributions. It is proposed by Onur C \etal~\cite{hamsici2007spherical}.

The definition of spherical-homoscedasticity resorts to the Gaussian approximation. We first give the definition of Gaussian approximation and then give the definition of spherical-homoscedasticity.

\noindent\textbf{Definition 1. }\emph{Suppose ${\bf x}_i$ is a sample from the spherical distribution. Then the Gaussian approximation is given as $N(E({\bf x}_i), Var(\bf{x}_i))$, where $E(.)$ and $Var(.)$ are the functions for expectation and variances.}

\noindent\textbf{Definition 2. }\emph{Suppose distribution $N_1(\mathbf{\bm \mu}, \Sigma)$ is the Gaussian approximation of spherical distribution $F_1$ and ${\bf A}$ is an orthogonal matrix. Suppose ${\bf A}$ is spanned by $\bm{\mu}$ and one of the eigenvectors of $\bm \Sigma$. Suppose $N_{2}\left(\mathbf{A} \bm{\mu}, \mathbf{A}^{T} \Sigma \mathbf{A}\right)$ is the Gaussian approximation of spherical distribution $F_2$. 
Then $N_1$ and $N_2$ ($F_1$ and $F_2$) are spherical-homoscedastic.}

Spherical-homoscedasticity requires the covariances of distributions to have identical eigenvalues. Geometrically, this property indicates that distributions share identical shape. In other words, distributions can be transformed to be totally overlapped.

\subsection{Spherical Feature Transform}
\label{sec:SFT}
Recently, \emph{feature normalization} has been widely discussed~\cite{ranjan2017l2L2face,wang2017normface,wang2018cosface} and adopted in DML frameworks~\cite{wang2019ranked,wang2019multi,wang2018cosface,deng2019arcface}. This technique scales all the features to the same norm. Thus, the features are restricted to lie on the surface of a hypersphere. In this case, the feature transform in Eq.~\ref{eq:translation} is no longer valid. There are two reasons. 
First, the identical-variance prior is too restrictive for spherical distributions. In general(e.g. the two distributions in Fig.~\ref{fig:illustration}(b)), spherical distributions are unlikely to have the same covariance. Second, \textit{translation transform} produces features lying out of the surface of the hypersphere. This breaks the manifold structure of the feature space as shown in Fig.~\ref{fig:illustration}(b). Therefore, both the identical-variance prior and the \emph{translation transform} should be modified for the spherical case.

\begin{figure}[!!t]
\begin{center}
\begin{tabular}{@{}c@{}c@{}c@{}}
\includegraphics[height=2.5cm]{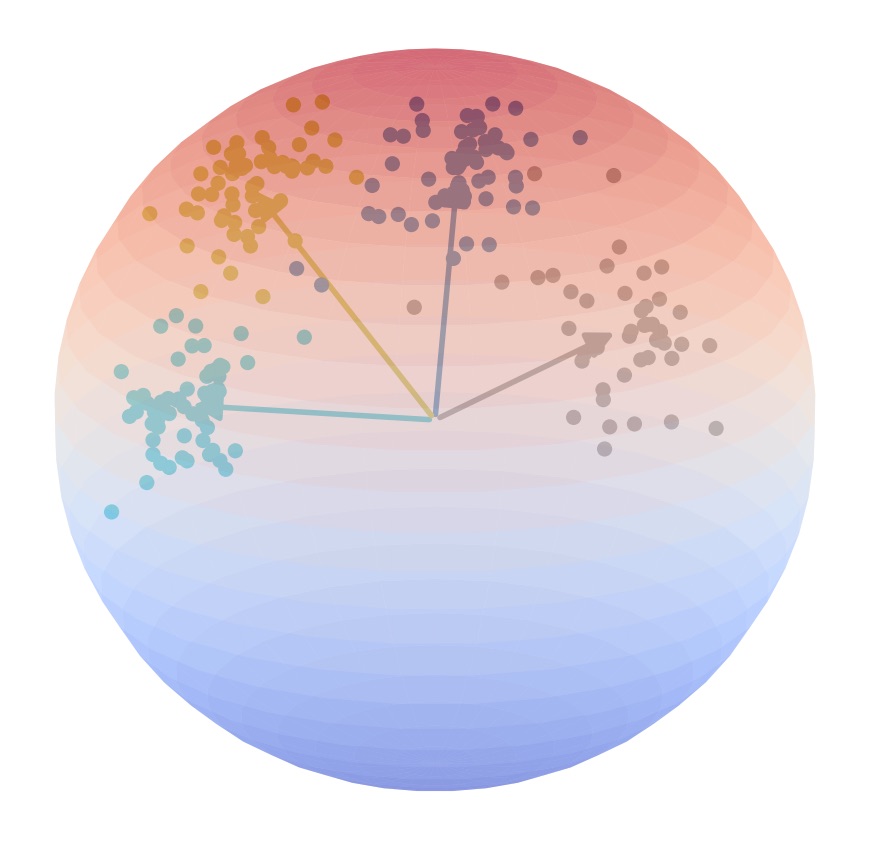} &
\includegraphics[height=2.5cm]{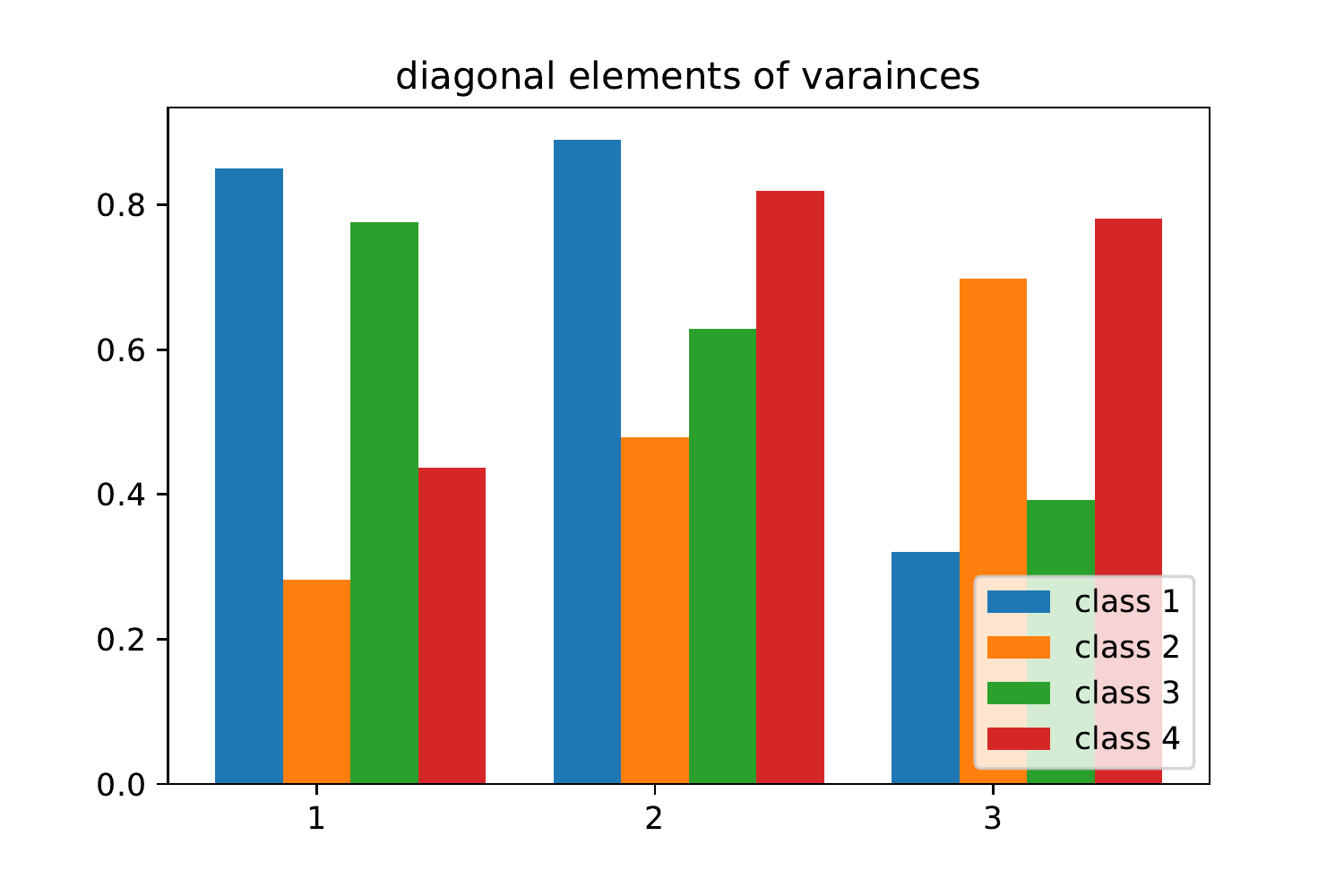} &
\includegraphics[height=2.5cm]{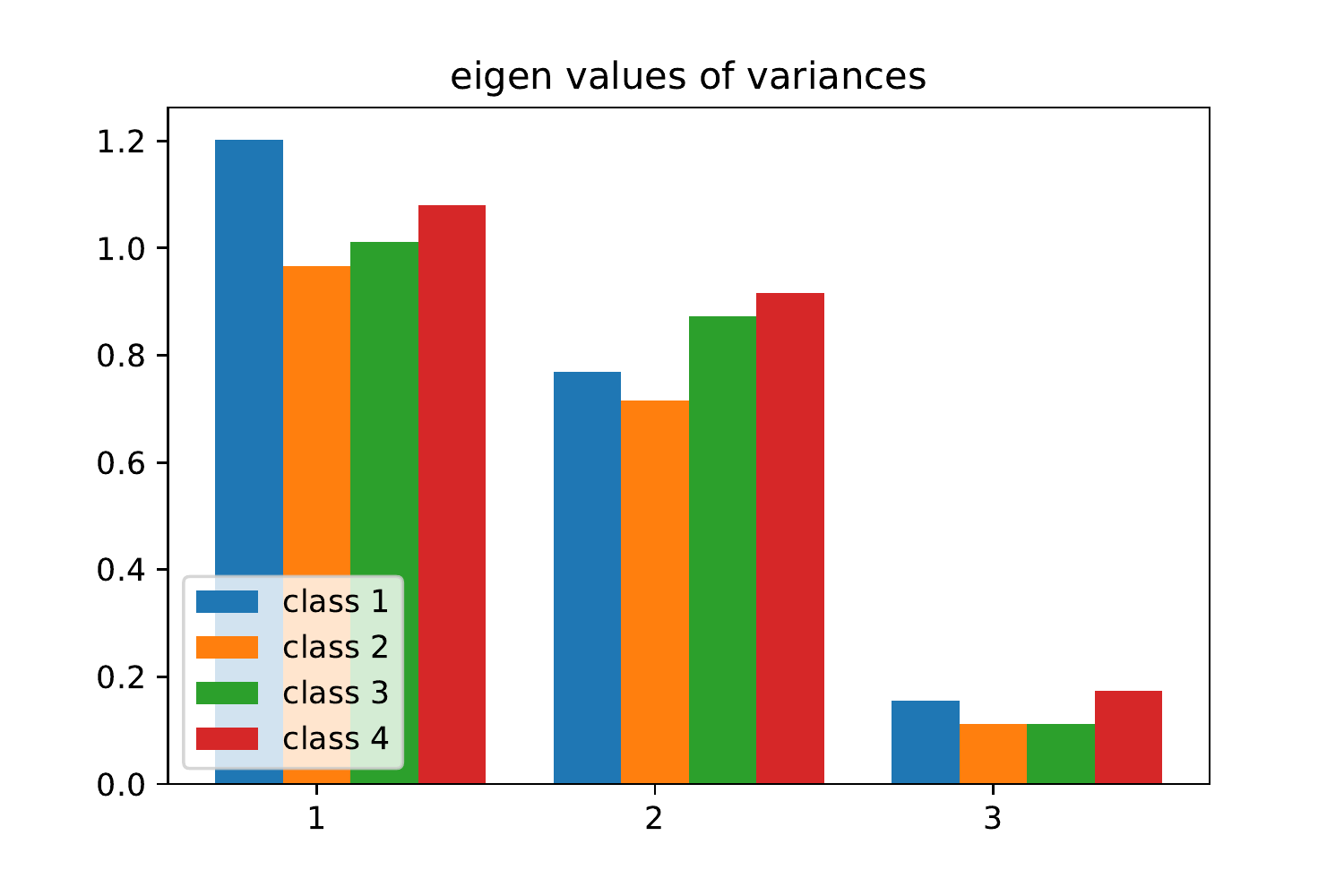}\\
\small (a) & (b) &(c)\\ [-1mm]
\end{tabular}
\end{center}
\caption{(a) Visualization of features on CUB dataset. Features are projected to 3D using PCA. (b) Diagonal elements of four classes' variances from CUB. The values from the same position on the diagonal are plotted together. They differ a lot. (c) Eigen values of four classes' variances from CUB. The eigen values from the same position on the eigen matrices are plotted together. They are much closer.}
\label{fig:simple_exp_of_sh}
\end{figure}

We propose a new approach. It relaxes the identical-variance prior to the prior of identical eigen values of variances, which is the spherical-homoscedasticity as defined in Sec.~\ref{sec:spherical}. This relaxation is validated Fig.~\ref{fig:simple_exp_of_sh}. The experiment is performed on CUB dataset (see experiments for details). We choose four classes with sufficient number of samples so that their feature distributions can be faithfully estimated. We compare their covariance matrices and the eigenvalues. As shown in Fig.~\ref{fig:simple_exp_of_sh}(b), their covariance matrices are significantly different, but the difference of the eigen values of these covariance matrices are much smaller (about 8\% on average) as shown in Fig.~\ref{fig:simple_exp_of_sh}(c). This shows that the identical-variance prior does not hold. And our assumption of identical eigenvalues of covariances is more valid. The similar observation is also found on other datasets in face recognition, vehicle recognition and etc. 

Geometrically, our assumpion implies that a distribution can be transformed to overlap with another via an orthogonal rotation matrix as in the 
Definition 2.
Thus, a feature vector in one class can be transformed to another class to generate augmented features. We denote the Gaussian approximation of two classes distributions as $N_1(\mathbf{\bm \mu_1}, \Sigma_1)$ and $N_2(\mathbf{\bm \mu_2}, \Sigma_2)$. Given a feature ${\bf x}_1$ sampled from $N_1$, we have:
\begin{equation}
{\bf \tilde{x}}_2 = {\bf Ax}_1,
\label{eq:rotate}
\end{equation} 
where ${\bf \tilde{x}}_2$ is considered as belonging to the class of $N_2$. This method is called \emph{Spherical Feature Transform~(SFT)}.

However, we note that solving the orthogonal matrix $\bf{A}$ according to 
Definition 2.
is non-trivial. A brute force approach would be complex. We propose a simpler and more elegant approach to calculate $\bf{A}$ without solving matrix equations. It is presented in the 
Proposition 1.\\
\textbf{Proposition 1. }\emph{Suppose $N_1(\mathbf{\bm \mu_1}, \Sigma_1)$ and $N_2(\mathbf{\bm \mu_2}, \Sigma_2)$ are two Gaussian approximations of spherical distributions. If they are spherical-homoscedastic, then the rotation matrix between them is spanned by $\bm{\mu}_1$ and $\bm{ \mu}_2$. }

The proof of 
Proposition 1
is left in the supplement. 
The rotation matrix $\bf{A}$ is calculated as following. First, we apply Schmidt orthogonalization to obtain $\bm{ \mu}_1$ and $\bm{\mu}_2$: ${\bf{n}}_1 = \bm{\mu}_1$, ${\bf n}_2 = \frac {\bm{\mu}_2 - (\bm{\mu}_2^T{\bf n}_1) {\bf n}_1}{\left\|\bm{\mu}_2 - (\bm{\mu}_2^T{\bf n}_1) {\bf n}_1\right\|_2}$. Then, we use Rodrigues rotation formula to calculate the rotation matrix: 

\begin{equation}
\begin{aligned}
{\bf A} = {\bf I} + ({\bf n}_2{\bf n}_1^T - {\bf n}_1{\bf n}_2^T)\sin(\alpha) + ({\bf n}_1{\bf n}_1^T + {\bf n}_2{\bf n}_2^T)(\cos(\alpha) - 1),
\label{eq:rodrigues}
\end{aligned}
\end{equation}
where ${\bf I}$ is the identity matrix and $\alpha$ is the rotation angle between $\bm{\mu}_1$ and $\bm{\mu}_2$. 

\subsection{Theoretical Analysis}
\label{sec:theo}

We discuss the relation between proposed SFT in Eq.(3) and the \emph{translation transform} in Eq.(2). In general, the two transforms are different. However, we show that a simple variant of the \emph{translation transform} (for normalized features) under some special cases is a degenerated form of SFT. Actually, we use this variant as a baseline method in our experiment.

In \emph{translation transform}, the variance part $\bm{\sigma}$ defined in Eq.(~\ref{eq:split}) is assumed to have the same distribution among all the classes. Differently, we propose SFT by showing that this term should be orthogonal transformed when features are normed. We observed that, when well trained, the features sampled from $\bm{\sigma}$ are likely to lie in the invariant subspace of the orthogonal matrix $\bf{A}$, as defined in 
Definition 2.
This observation is experimentally validated in Sec~\ref{sec:exp_degeneration}. We show that in this case SFT degenerates to \textit{translation transform} defined in Eq.~\ref{eq:translation}. With this condition $\bf{A}\bm{\sigma}_1 = \bm{\sigma}_1$, Eq.~(\ref{eq:rotate}) is simplified as

\begin{equation}
\begin{aligned}
{\bf \tilde{x}}_2 = {\bf Ax}_1 = {\bf A}(\bm{\mu}_1 + \bm{\sigma}_1) = \bm{\mu}_2 + \bm{\sigma}_1.
\label{eq:simple_degeneration}
\end{aligned}
\end{equation}

The right side is the \emph{translation transform} in Eq.(2).

\begin{figure}[!!t]
\begin{center}
\begin{tabular}{@{}c@{\hspace{2mm}}c@{}}
\includegraphics[height=2.5cm]{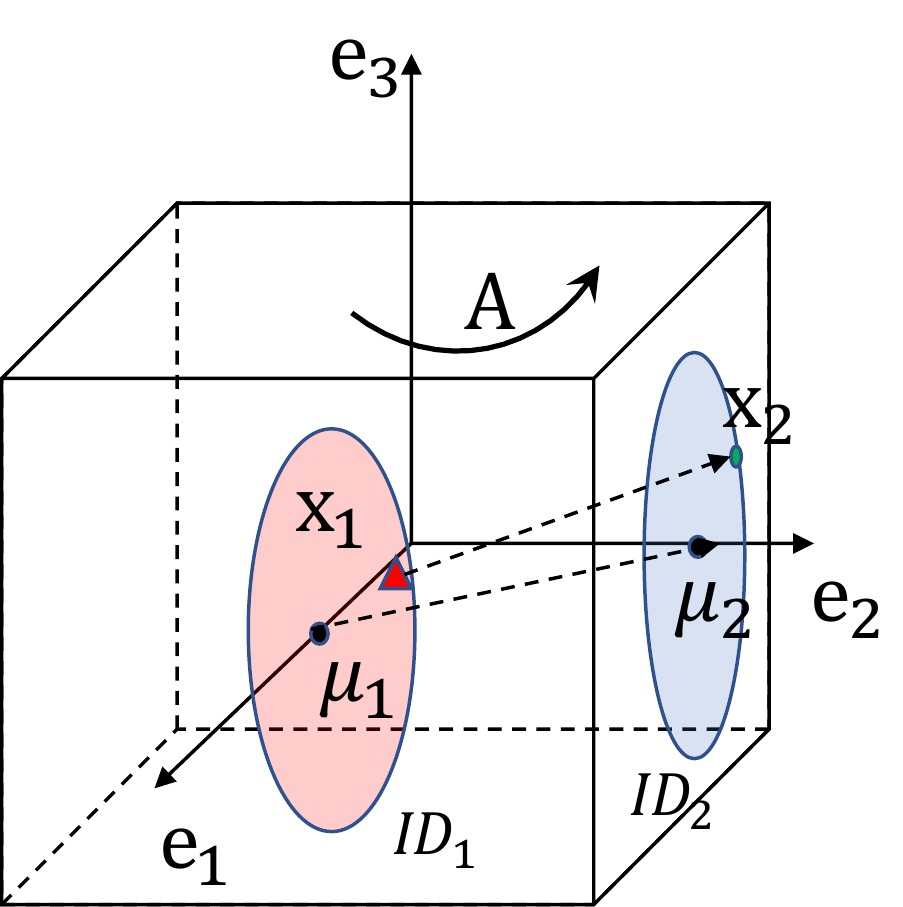} &
\includegraphics[height=2.5cm]{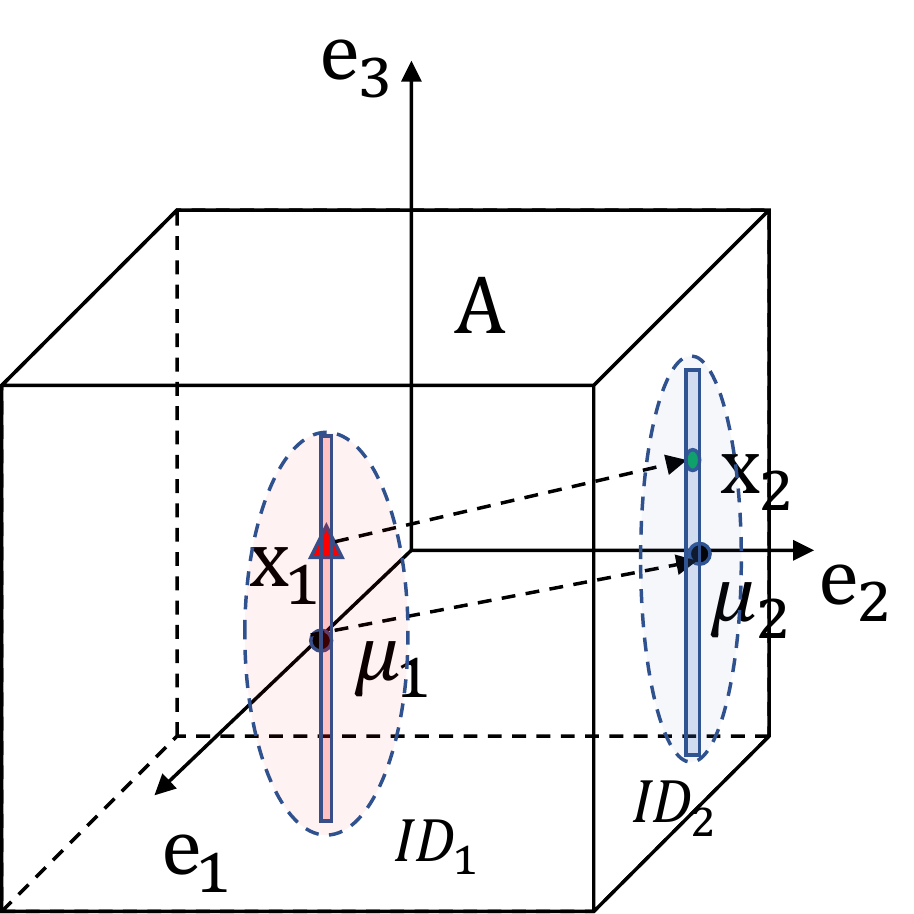} \\
\small (a) & (b) \\ [-1mm]
\end{tabular}
\end{center}
\caption{
Illustration of the degeneration from SFT to \emph{translation transform} by taking a three-dimensional example. The ${\bf e}_1$, ${\bf e}_2$, ${\bf e}_3$ are three axis of the coordinate. The red and blue ellipses represents the distribution of $\text{ID}_1$ and $\text{ID}_2$. Suppose they are spherical-homoscedastic and the rotation matrix between them is $\bf A$ and ${\bf x}_2 = {\bf Ax}_1$ (a) In general $\bm \mu_2 - \bm \mu_1$ is not equivalent to ${\bf x}_2 - {\bf x}_1$. 
(b) Special case: The intra-class variances are now encoded by the one-dimensional space spanned by ${{\bf e}_3}$ and the $\bm \mu_2 - \bm \mu_1$ is equivalent to ${\bf x}_2 - {\bf x}_1$.
}
\label{fig:illustration_2}
\end{figure}

This degeneration case is a bit hard to understand, especially in high dimensional space. For an intuitive illustration, we show an example in three-dimensional space. As shown in Fig.~\ref{fig:illustration_2}(a), in general, the result of SFT ${\bf x}_2 = {\bf Ax}_1$ is not equal to ${\bf x}_1 + \bm{\mu}_2 - \bm{\mu}_1$. While, some special features stay equal after rotation and translation, as shown in Fig.~\ref{fig:illustration_2}(b). In such a case, the direction of $\bm{\sigma}_1$ is parallel to the rotation axis of $\bf{A}$. That is, $\bm{\sigma}_1$ lie in the invariant subspace of $\bf{A}$.

\noindent\textbf{Proposition 2. }\emph{The degeneration happens only before feature normalization.}
\textit{Proof} Suppose feature $\bf{x} = \bm{\mu}+\bm{\sigma}$ and its variance part $\bm{\sigma}$ lie in the invariant subspace of $\bf{A}$. As $\bf{A}$ is spanned by $\bm{\mu}$ and one of the other vector, $\bm{\mu}$ is orthogonal to the invariant subspace of $\bf{A}$. So $\bm{\mu}$ is orthogonal to $\sigma$. Then the norm of $\bf{x}$ is evaluated as:
\begin{equation}
\begin{aligned}
\|\bf{x}\| = \|\bm{\mu}+\bm{\sigma}\| = \sqrt{(\bm{\mu}+\bm{\sigma})^T(\bm{\mu}+\bm{\sigma})} = \sqrt{\bm{\mu}^T\bm{\mu}+\bm{\sigma}^T\bm{\sigma}}
\label{eq:scatter_def}
\end{aligned}
\end{equation}
As $\bm{\mu}$ is a constant for one class and the $\bm{\sigma}$ varies, the norm of $\bf{x}$ can not be a constant for each features of a class. In other words, the feature norms are not constant. \hfill $\square$

Based on 
Proposition 2
, we can make a simple modification to the \emph{translation transform} to make it able to produce valid features in spherical case. Specifically, we use Eq.~\ref{eq:translation} before feature normalization and then reproject them back to the hypersphere. This variant is referred to as the degenerated form of SFT. 

However, the degenerated form will produce identical augmented features as SFT only when degeneration takes place. There are still features that won't obey the condition of degeneration. Directly applying the degenerated form on them may curse the augmentation process. 
We further investigate into whether there is an ideal case where the degeneration will always take place thus the degenerated form can be treated as an alternative of SFT. Considering the condition of degeneration, this special case should satisfy $\bf{A}\bm{\sigma} = \bm{\sigma}$ for any rotation matrix $\bf{A}$ and any $\bm{\sigma}$. The three-dimensional example in Fig.~\ref{fig:illustration_2}(b) gives a clear clue that this special case exist mathematically. Specifically, if the feature distributions are shrunk in the plane defined by $\{\bf{e_1}, \bf{e_2}\}$, then all $\bm{\sigma}$ will lie in the invariant subspace of $\bf{A}$. The exact mathematical description for such case is presented in 
Proposition 3\\
\noindent\textbf{Proposition 3. }\emph{SFT degenerates to the \emph{translation transform} iff for $\forall$ feature ${\bf x}$ with ${\bf x} = \bm{\mu}+\bm{\sigma}$, $\bm{\mu} \perp \bm{\sigma}$}

The proof is presented in the supplement. 
Proposition 3
has revealed a extremely restrictive condition that the mean vectors $\bm{\mu}$ and $\bm{\sigma}$ lie in two orthogonal subspaces. Intuitively, this condition is hard to be satisfied. While surprisingly, it is found, although not clear why, but in general, that deep neural networks tend to learn an orthogonal subspaces for $\bm{\mu}$ and $\bm{\sigma}$. For revealing this phenomenon, we define a measure of how much the the subspaces of $\bm{\mu}$ and $\bm{\sigma}$ are orthogonal. We first define two covariance matrices:
\begin{equation}
\begin{aligned}
{\bf S}_c = \frac{1}{C}\sum_{i=1}^{C}{\bm \mu_i}^T{\bm \mu_i}, 
{\bf S}_w = \frac{1}{C}\sum_{i=1}^{C}\frac{1}{N_k}\sum_{k=0}^{N_k}\left({\bf x}_k - {\bm \mu}_i\right)^T\left({\bf x}_k - {\bm \mu}_i\right),
\label{eq:scatter_def}
\end{aligned}
\end{equation}
where $y_k$ is the label of embedding $\bf x_k$. $C$ is the number of classes. $N_k$ is the number of samples for $k$-th class.
Then, we estimate the eigenvalue space for ${\bf S}_c$ and denote them as ${\bf U}$, where ${\bf U} = \{\bf u_1, u_2,..., u_k\}$ corresponds to the $k$ largest eigenvalues. The subspace spanned by ${\bf U}$ will cover most energy of the mean vectors while the energy of $\sigma$ will distribute over these components. We calculate the remaining energy percent of $\sigma$ in this subspace by evaluating:
\begin{equation}
\begin{aligned}
r_w = trace\left({\bf U}^T {\bf S}_w {\bf U}\right)/trace({\bf S}_w),
\end{aligned}
\label{eq:degeneration_ratio}
\end{equation}
where $trace(.)$ is the sum of the diagonal elements. $r_w$ measures that how much energy percent of $\bm{\sigma}$ is distributed over the subspace spanned by ${\bf U}$. 
$r_w$ is between 0 and 1. If $r_w=0$, then the subspaces for $\bm{\mu}$ and $\bm{\sigma}$ are orthogonal. So the smaller $r_w$, the nearer of the state being orthogonal.

\subsection{Training Scheme}
\label{sec:training}
Both the \textit{translation transform} defined in Eq.~\ref{eq:translation} and SFT defined in Eq.~\ref{eq:rotate} rely on the accurate estimation of the feature center of each class. 
We denote the feature centers as  $\{\bm{\mu}_1, \bm{\mu}_2, ...,\bm{\mu}_C\}$ where $C$ is the number of classes. In every mini-batch, we update them by:
\begin{equation}
\Delta {\bm \mu}_{j}=\frac{\sum_{i=1}^{m} \delta\left(y_{i}=j\right) \cdot\left({\bm \mu}_{j}-{\bf x}_{i}\right)}{1+\sum_{i=1}^{m} \delta\left(y_{i}=j\right)},
\label{eq:center_update}
\end{equation}
where $y_i$ is the label of feature ${\bf x}_i$ and $m$ is the mini-batch size. $\delta(.)$ is the indicator function.
For training, we propose two train schemes depending on the whether the training set is balanced. \\
\noindent{\bf{Balanced train. }} When the dataset is balanced in the number of samples for each class, we will generate new features for every class. In specific, for a feature, we randomly choose a different class as target and transform the feature to that class. We do this for every feature in the mini-batch. After that, we get a new batch of features with different labels.\\
\noindent{\bf{Unbalanced train. }}  When the dataset is unbalanced in the number of samples for each class, we only generate new features for classes that are short of samples. In specific, we set a threshold for the number of samples and use it to separate the whole training data into head classes and tail classes. For any head features in a mini-batch, we randomly choose a tail class as target and transform it into the tail distributions. 

For both training schemes, we get two batch of training data. Let $\mathbf{X}=\left[\mathbf{x}_{1}, \cdots, \mathbf{x}_{n}\right]$ be the original features and $\mathbf{Y}=\left[y_{1}, \cdots, y_{n}\right]$ be the corresponding labels, where $y_{i} \in\{1, \cdots, C\}$. Let $\mathbf{X}_{gen}=\left[\mathbf{x}_{gen,1}, \cdots, \mathbf{x}_{gen,m}\right]$ be the generated batch and $\mathbf{Y}_{gen}=\left[y_{gen,1}, \cdots, y_{gen,m}\right]$ be the corresponding labels. As our augmentation method is applicable to any DML frameworks, we denote $J(\bm \theta; \mathbf{X}, \mathbf{Y})$ as a general target function with $\bm \theta$ denoting the parameters to be optimized and $\mathbf{X}$, $\mathbf{Y}$ denoting the batch data and labels. Similar to DVML~\cite{lin2018deepVariational} and HDML~\cite{zheng2019hardnessHDML}, we also apply the metric learning losses on the original features besides the augmented features. It is because that the augmentation process relies on a well trained feature space. Omitting the original features or applying too much weight on the augmented features will curse the training process. It is shown in Sec~\ref{sec:ablation}. We formulate our losses as:
\begin{equation}
\min _{\bm \theta} J= J(\bm \theta; \mathbf{X}, \mathbf{Y}) + \lambda J(\bm \theta; \mathbf{X}_{gen}, \mathbf{Y}_{gen}),
\label{eq:training_target}
\end{equation}
where $\lambda$ is a weighting factor controlling the balance between the original batch data and the generated batch data. The total training scheme for feature transform is illustrated in Algorithm~\ref{algorithm:training_scheme}.

\begin{algorithm}[!ht]
    \begin{algorithmic}[1]
        \Require Training image set, network $f$, target function $J$, parameters $\lambda$ and number of iteration numbers T.
        \Ensure Parameters of network $\bm \theta$
        \State Initialize $\bm \theta$
        \For{$iter = 1, ..., T$}
        \State Sample mini-batch of $m$ training images.
        \State Extract embeddings using $f$ to get $\bf X$ with labels $\bf Y$.
        \State Produce data $\{{\bf X_{gen}}, {\bf Y_{gen}}\}$ using (\ref{eq:rotate}) or (\ref{eq:translation}).
        \State Update geometric centers using (\ref{eq:center_update}).
        \State Optimize $\bm \theta$ using (\ref{eq:training_target}).
        \EndFor
    \end{algorithmic}
    \caption{Training with Feature Transform}
    \label{algorithm:training_scheme}
\end{algorithm}

\section{Experiments}
\label{sec:exp}
\paragraph{\textbf{Datasets and Metrics.}}
We conduct experiments on two types of benchmark datasets: Metric Learning and Face Recognition. 
For metric learning, we experiment on three widely-used benchmarks to evaluate the our approach: 
(1){\bf Cars196}~\cite{krause20133dCARS}, (2){\bf CUB-200-2011}~\cite{wah2011caltechCUB}, (3)Stanford Online Products~({\bf SOP})~\cite{oh2016deepLifted}.
To evaluate the performance of each method, we follow \cite{duan2018deepAdversarial} to perform the K-means algorithm in the test set and report normalized mutual information (NMI) and $F_1$ metrics as well as Recall@K for retrieval task. 
For face recognition, we use a cleaned version of MS-Celeb-1M~\cite{guo2016ms1m} as our training set that contains 3M facial images and 80920 classes. We present evaluation results on three face verification benchmarks: {\bf LFW}~\cite{huang2008labeledLFW}, {\bf YTF}~\cite{wolf2011faceYTF} and {\bf IJB-C}~\cite{maze2018iarpaIJB-C}. For LFW and YTF, we follow the unrestricted with labeled outside data protocol and report the performance of 6,000 face pairs on LFW and 5,000 video pairs on YTF. For IJB-C, we follow the 1:1 verification protocol to evaluate 19,557 positive matches and 15,638,932 negative matches and report the results of TARs at various FARs.

\paragraph{\noindent\textbf{Implementation Details.}}
For the metric learning task, we use GoogleNet~\cite{googlenet}(or GoogleNet-V2) pre-trained with ImageNet~\cite{imagenet_cvpr09} as a backbone network and add a fully connected layer at the end to output the feature embedding. We use the same data preprocessing and augmentation as in Multi-Similarity Loss~~\cite{wang2019multi}. 
We set the embedding size to 512 and perform $\ell_2$-normalization on the feature.
We use the SGD optimizer with a weight decay of 1e-4 and train for 30,000 iterations.
For learning rate, we set 1e-2 for Cars196 and SOP and 1e-3 for CUB-200-2011 as base learning rate for backbone and newly added layers 10x the base learning rate, and decay the learning rate by multiply 0.1 every 10,000 iterations. 
We set the batch size to be 60 made up of 20 classes and 3 images per class. The balanced train scheme is adopted when SFT is used.
For face recognition, the CNN architecture used in our work is similar to \cite{liu2017sphereface}. We change the number of residual units to $[3,4,6,3]$ to construct a 34-layer residual network. We preprocess all face images by MTCNN~\cite{zhang2016jointMTCNN}. Then the 5 facial points are adopted to perform alignment to the face image. After that, we resize the cropped image to $112 \times 112$. Each pixel(in [0, 255]) in RGB images is normalized by subtracting 127.5 then being divided by 128. We use SGD optimizer with a weight decay of 5e-4 and train for 120K iterations. The learning rate is set to 0.1 initially and is divided by 10 at the 70K, 90K and 110K iterations. The unbalanced train scheme is adopted when SFT is used, where we set the classes that have less than 15 samples as tail classes.

\begin{table*}[!t]
\caption{Comparison on Cars196 and CUB-200-2011.}
\label{table:sota_car_cub}
\centering
\resizebox{\linewidth}{!}{
\setlength{\tabcolsep}{9pt}
\begin{tabular}{lcccccccccc}
\hline
 & \multicolumn{5}{c|}{Cars196} & \multicolumn{5}{c}{CUB-200-2011} \\ \cline{2-11} 
\textbf{} & R@1 & R@2 & R@4 & MNI & \multicolumn{1}{c|}{F1} & R@1 & R@2 & R@4 & NMI & F1 \\ \hline
\multicolumn{11}{c}{GoogleNet} \\ \hline
Triplet & 58.4 & 70.3 & 80.2 & 57.0 & \multicolumn{1}{c|}{27.2} & 42.8 & 55.2 & 55.6 & 52.4 & 19.1 \\
Triplet+HDML~\cite{zheng2019hardnessHDML} & 62.0 & 73.3 & 82.9 & 57.7 & \multicolumn{1}{c|}{27.8} & 44.3 & 56.0 & 68.0 & 55.5 & 26.7 \\
Triplet+DVML~\cite{lin2018deepVariational} & 64.4 & 73.5 & 78.6 & \textbf{60.5} & \multicolumn{1}{c|}{28.4} & 43.3 & 55.8 & 68.0 & 55.0 & 25.2 \\
Triplet+FTL & 60.1 & 71.5 & 80.5 & 57.9 & \multicolumn{1}{c|}{25.0} & 46.8 & 59.2 & 70.2 & 57.3 & 24.3 \\
Triplet+SFT-d & 60.3 & 71.7 & 81.4 & 57.9 & \multicolumn{1}{c|}{28.1} & 46.5 & 59.3 & 70.0 & 57.9 & 28.1 \\
Triplet+SFT & \textbf{65.1} & \textbf{75.7} & \textbf{84.0} & 58.1 & \multicolumn{1}{c|}{\textbf{28.6}} & \textbf{48.3} & \textbf{60.0} & \textbf{71.2} & \textbf{58.1} & \textbf{28.6} \\ \hline
NPair & 72.8 & 82.3 & 88.5 & 61.3 & \multicolumn{1}{c|}{29.4} & 53.5 & 64.9 & 72.3 & 60.4 & 27.8 \\
NPair+HDML~\cite{zheng2019hardnessHDML} & 78.9 & 87.0 & 91.0 & 67.1 & \multicolumn{1}{c|}{37.3} & 53.9 & 65.8 & 76.7 & 62.0 & 30.0 \\
NPair+DVML~\cite{lin2018deepVariational} & \textbf{80.2} & 85.6 & 91.9 & 66.1 & \multicolumn{1}{c|}{34.8} & 54.2 & 66.2 & 77.3 & 62.0 & 31.5 \\
NPair+FTL & 73.1 & 82.2 & 88.6 & 60.0 & \multicolumn{1}{c|}{27.4} & 54.0 & 66.0 & 77.0 & 61.9 & 29.7 \\
NPair+SFT-d & 76.2 & 85.0 & 90.9 & 64.2 & \multicolumn{1}{c|}{33.1} & 54.5 & 67.0 & \textbf{77.7} & 62.0 & 30.1 \\
NPair+SFT & 79.4 & \textbf{87.1} & \textbf{92.4} & \textbf{67.2} & \multicolumn{1}{c|}{\textbf{37.3}} & \textbf{54.7} & \textbf{67.0} & 77.5 & \textbf{62.2} & \textbf{30.5} \\ \hline
\multicolumn{11}{c}{GoogleNet-V2} \\ \hline
RLL~\cite{wang2019ranked} & 74.2 & 83.2 & 89.0 & 62.2 & \multicolumn{1}{c|}{32.9} & 59.6 & 71.0 & 80.5 & 64.3 & 32.9 \\
RLL+DVML~\cite{lin2018deepVariational} & 79.0 & 86.6 & 91.3 & 65.5 & \multicolumn{1}{c|}{34.9} & 60.2 & 71.7 & 81.0 & 64.7 & 33.0 \\
RLL + SFT-d & 78.8 & 86.7 & 92.1 & 65.4 & \multicolumn{1}{c|}{34.4} & 59.4 & 71.2 & 80.9 & 64.2 & 32.8 \\
RLL + SFT & \textbf{80.2} & \textbf{88.1} & \textbf{92.8} & \textbf{66.1} & \multicolumn{1}{c|}{\textbf{35.3}} & \textbf{60.3} & \textbf{71.8} & \textbf{81.1} & \textbf{64.9} & \textbf{33.6} \\ \hline
MS~\cite{wang2019multi} & 84.0 & 90.2 & 94.1 & 72.8 & \multicolumn{1}{c|}{45.3} & 65.7 & 76.6 & 84.6 & 69.0 & 39.6 \\
MS+DVML~\cite{lin2018deepVariational} & 84.4 & \textbf{90.8} & 92.4 & 72.0 & \multicolumn{1}{c|}{45.3} & 66.2 & 76.7 & 85.1 & 69.6 & 40.0 \\
MS + SFT-d & 83.8 & 90.4 & 94.6 & 73.1 & \multicolumn{1}{c|}{45.3} & 66.1 & 76.8 & 85.2 & 70.0 & \textbf{41.6} \\
MS + SFT & \textbf{84.5} & 90.6 & \textbf{94.6} & \textbf{73.2} & \multicolumn{1}{c|}{\textbf{45.8}} & \textbf{66.8} & \textbf{77.5} & \textbf{85.8} & \textbf{70.3} & 40.4 \\ \hline
\end{tabular}
}
\end{table*}

\paragraph{\textbf{Compared Methods.}}
We compare our method to other feature generation methods, including HDML~\cite{zheng2019hardnessHDML}, DVML~\cite{lin2018deepVariational} and FTL~\cite{featureTransferLearning}. These methods are introduced in Sec~\ref{sec:related}. They require no extra labeling and can be compared fairly on metric learning tasks. Also the degenerated SFT will be included for comparison. It is denoted as SFT-d in the results.
The comparison is made on two traditional representative baseline losses, aka, triplet loss~\cite{schroff2015facenet} and NPair loss~\cite{sohn2016npair} and two most recent baseline losses that achieved high results, aka, Ranked List Loss~(RLL)~\cite{wang2019ranked} and Multi-Similarity Loss~(MS)~\cite{wang2019multi}. Most of the comparison is made on GoogleNet~\cite{googlenet} because almost all of the chosen competitors report their results on this backbone. For comparison with the SOTA, we also make some comparison on GoogleNet-V2. 
For fair comparison, we implement all of these methods and report the results from our experiments. 

For FTL~\cite{featureTransferLearning}, the features are normed in our implementation as we found that feature normalization will outperform the original method greatly. The FTL differs from the degenerated form of SFT in that it requires a pre-training of the network and is applied in the fine-tuning stage while this is not needed in both SFT and the degenerated form. Also, FTL requires a decoder network and only transfers a part of the energy of $\bm{\sigma}$ using PCA. In our implementation, we follows them to use 95\%.

\subsection{Quantitative Results}
Table~\ref{table:sota_car_cub}, Table~\ref{tab:sop}, Table~\ref{tab:lfw_ytf} and Table~\ref{tab:ijb_c} present the experimental results of SFT on three popular deep metric learning benchmarks and three face recognition benchmarks respectively. 

By comparing with baseline methods, it is noticed that SFT can significantly improve the performance of them, especially on Cars196 and CUB-200-2011. For example, when coupled with NPair loss, SFT improves the baseline by 7 point on Cars196. 
SFT can also boost performance on higher baselines that reported by two most recent losses, Multi-Similarity loss and Ranked-List loss. 
While SFT is relatively less effective on SOP(Table~\ref{tab:sop}). The reason is that the number of samples for each class in SOP is too small(about 5).

To sum up, SFT performs better than HDML~\cite{zheng2019hardnessHDML}, FTL~\cite{featureTransferLearning} and DVML~\cite{lin2018deepVariational}. For example, when coupled with NPair loss, SFT outperforms the HDML by 1.0 on Cars196. 
\begin{wraptable}{r}{0.5\textwidth}
\caption{Face verification (\%) on the LFW and YTF datasets.}
\label{tab:lfw_ytf}
\centering
\resizebox{\linewidth}{!}{
\begin{tabular}{lccc}
\hline
Method                                   & Training Data & LFW                       & \multicolumn{1}{l}{YTF}  \\ \hline
DeepFace+~\cite{taigman2014deepfaceplus} & 4M            & 97.35                     & 91.4                     \\
FaceNet~\cite{schroff2015facenet}        & 200M          & 99.63                     & 95.1                     \\
DeepID2+~\cite{sun2015deeplyDeepID2}     & 300K          & 99.47                     & 93.2                     \\
SphereFace~\cite{liu2017sphereface}      & 0.5M          & 99.42                     & 95.0                     \\
CosFace~\cite{wang2018cosface}           & 5M            & 99.73                     & 97.6                     \\
ArcFace~\cite{deng2019arcface}           & 5.8M          & 99.83                     & 98.02                    \\
L2-Face~\cite{ranjan2017l2L2face}        & 3.7M          & 99.78                     & 96.08                    \\ \hline
L2-Face~\cite{ranjan2017l2L2face} (ours)  & 3M            & \multicolumn{1}{l}{99.45} & \multicolumn{1}{l}{96.0} \\
L2-Face(ours) + SFT-d                        & 3M            & \multicolumn{1}{l}{99.41} & \multicolumn{1}{l}{95.9} \\
L2-Face(ours) + SFT                        & 3M            & \multicolumn{1}{l}{99.50} & \multicolumn{1}{l}{96.5} \\ \hline
CosFace~\cite{wang2018cosface} (ours)    & 3M            & 99.68                     & 96.2                     \\
CosFace (ours) + SFT-d                       & 3M            & 99.70                     & 96.5                     \\
CosFace (ours) + SFT                       & 3M            & 99.73                     & 97.2                     \\ \hline
\end{tabular}
}
\end{wraptable}
On higher baselines, such as Multi-Similarity Loss, our SFT outperforms DVML by 0.7 on CUB. The degenerated form of SFT can be effective on most baseline methods. While averagely, it surpass the performance of SFT.
Besides the metric learning losses, SFT can also be used together with softmax-based losses. This is mainly used in face recognition tasks. On LFW dataset and YTF dataset(Shown in Table~\ref{tab:lfw_ytf}), the performance of deep neural networks are nearly saturated, but we still report the performance for comparison with the other works. On IJB-C(Shown in Table~\ref{tab:ijb_c}), we provide a competitive baseline for both L2-Face~\cite{ranjan2017l2L2face} and CosFace~\cite{wang2018cosface}, while it is observed that SFT can still boost the performance when compared with the baselines.

\begin{table}[!t]
\begin{minipage}{.45\textwidth}
\caption{Experimental results on Stanford Online Products(SOP). SFT is less effective on SOP as the number of samples for each class is only 5.}
\label{tab:sop}
\centering
\resizebox{!}{1.7cm}{
\begin{tabular}{lccccc}
\hline
 & \multicolumn{5}{c}{SOP} \\ \cline{2-6} 
 & R@1 & R@10 & R@100 & NMI & F1 \\ \hline
Triplet (ours) & 70.8 & 85.5 & 93.8  & 88.2 & 28.0 \\
Triplet + SFT-d                               & 71.9 & 86.4 & 94.4  & 88.5 & 29.3 \\
Triplet + SFT                                & 72.3 & 86.5 & 94.5  & 88.6 & 29.9 \\ \hline
RLL~\cite{wang2019ranked} (ours) & 77.5 & 89.9 & 95.8 & 89.7 & 35.3 \\
RLL + SFT-d & 77.9 & 90.3 & 96.1 & 89.8 & 35.9 \\
RLL + SFT & 77.8 & 90.2 & 96.0 & 89.9 & 36.4 \\ \hline
MS~\cite{wang2019multi} (ours) & 73.1 & 87.2 & 94.7 & 88.5 & 29.6 \\
MS + SFT-d & 73.5 & 87.5 & 94.9 & 88.6 & 29.8 \\
MS + SFT & 73.4 & 87.1 & 94.7 & 88.8 & 30.9 \\ \hline
\end{tabular}
}
\end{minipage}
\begin{minipage}{.5\textwidth}
\caption{Comparison with the state-of-art methods on IJB-C. The `-' denotes the corresponding results are not reported in the original paper.}
\label{tab:ijb_c}
\centering
\resizebox{!}{1.7cm}{
\begin{tabular}{lcccc}
\hline
\multirow{2}{*}{Method}                 & \multirow{2}{*}{Training Data} & \multicolumn{3}{c}{IJB-C(TAR@FAR)}                                                   \\ \cline{3-5} 
                                        &                                & \multicolumn{1}{l}{0.001\%} & \multicolumn{1}{l}{0.01\%} & \multicolumn{1}{l}{0.1\%} \\ \hline
Vggface2~\cite{cao2018vggface2}         & 3.3M                           & 74.7                        & 84.0                       & 91.0                      \\
L2-Face~\cite{ranjan2017l2L2face}       & 3.3M                           & 78.54                       & 87.01                      & 92.10                     \\
Arcface~\cite{deng2019arcface}          & 5.8M                           & -                           & 92.10                      & -                         \\ \hline
L2-Face~\cite{ranjan2017l2L2face}(ours) & 3M                             & 79.3                        & 87.3                       & 93.3                      \\
L2-Face(ours) + SFT-d                   & 3M                             & 79.4                        & 87.9                       & 93.3                      \\
L2-Face(ours) + SFT                       & 3M                             & {\bf 80.6}                  & {\bf 88.2}                 & {\bf 93.6}                \\ \hline
CosFace~\cite{wang2018cosface} (ours)   & 3M                             & 85.67                       & 92.11                      & 95.4                      \\
CosFace (ours) + SFT-d                   & 3M                             & 86.85                       & {\bf 92.78}                & {\bf 95.72}               \\
CosFace (ours) + SFT                      & 3M                             & {\bf 87.19}                 & 92.63                      & 95.6                      \\ \hline
\end{tabular}
}
\end{minipage}
\end{table}

\subsection{Ablation Study}
\label{sec:ablation}
In this part, we conduct the ablation study on Cars-196 with the ranked list loss. The conclusions from these experiments are also applicable to other datasets and loss functions.

\begin{figure}[!t]
\centering
\begin{tabular}{@{}c@{}c@{}c@{}c@{}}
\includegraphics[width=0.4\textwidth]{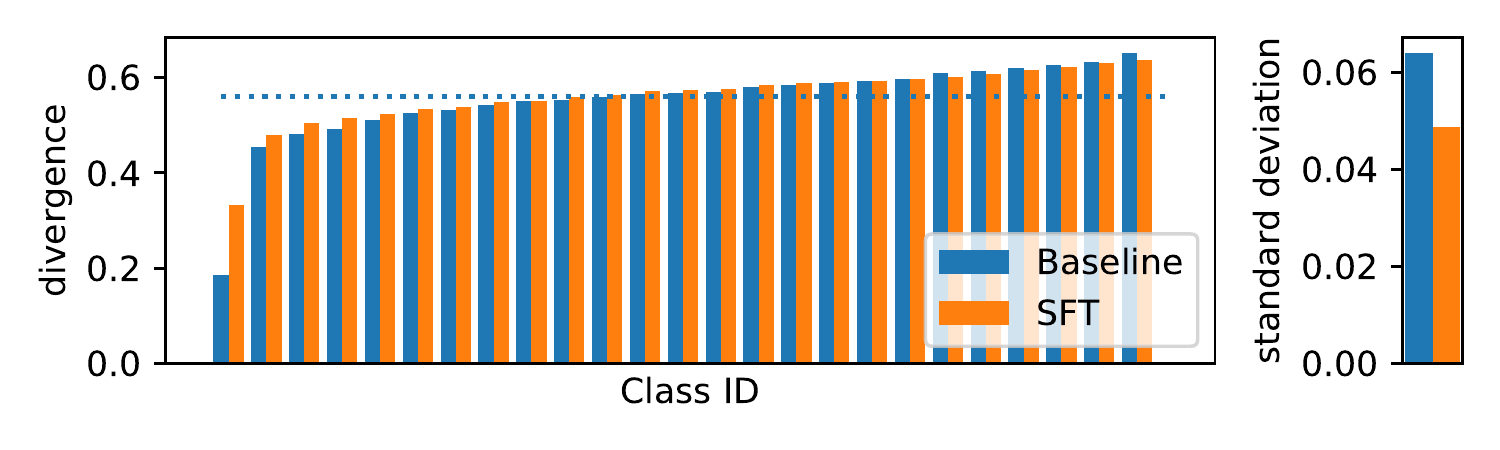} &
\includegraphics[width=0.2\textwidth]{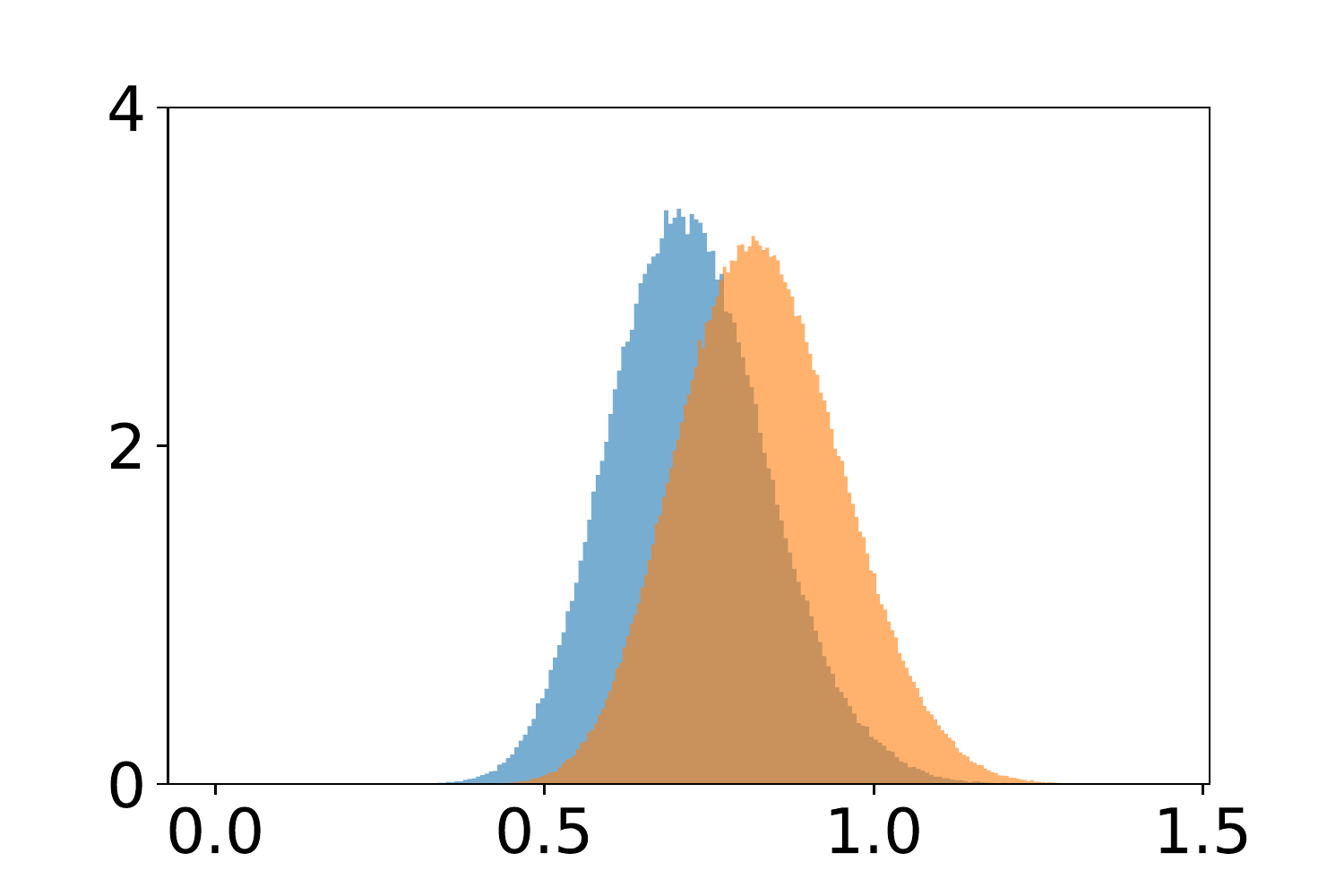} &
\includegraphics[width=0.2\textwidth]{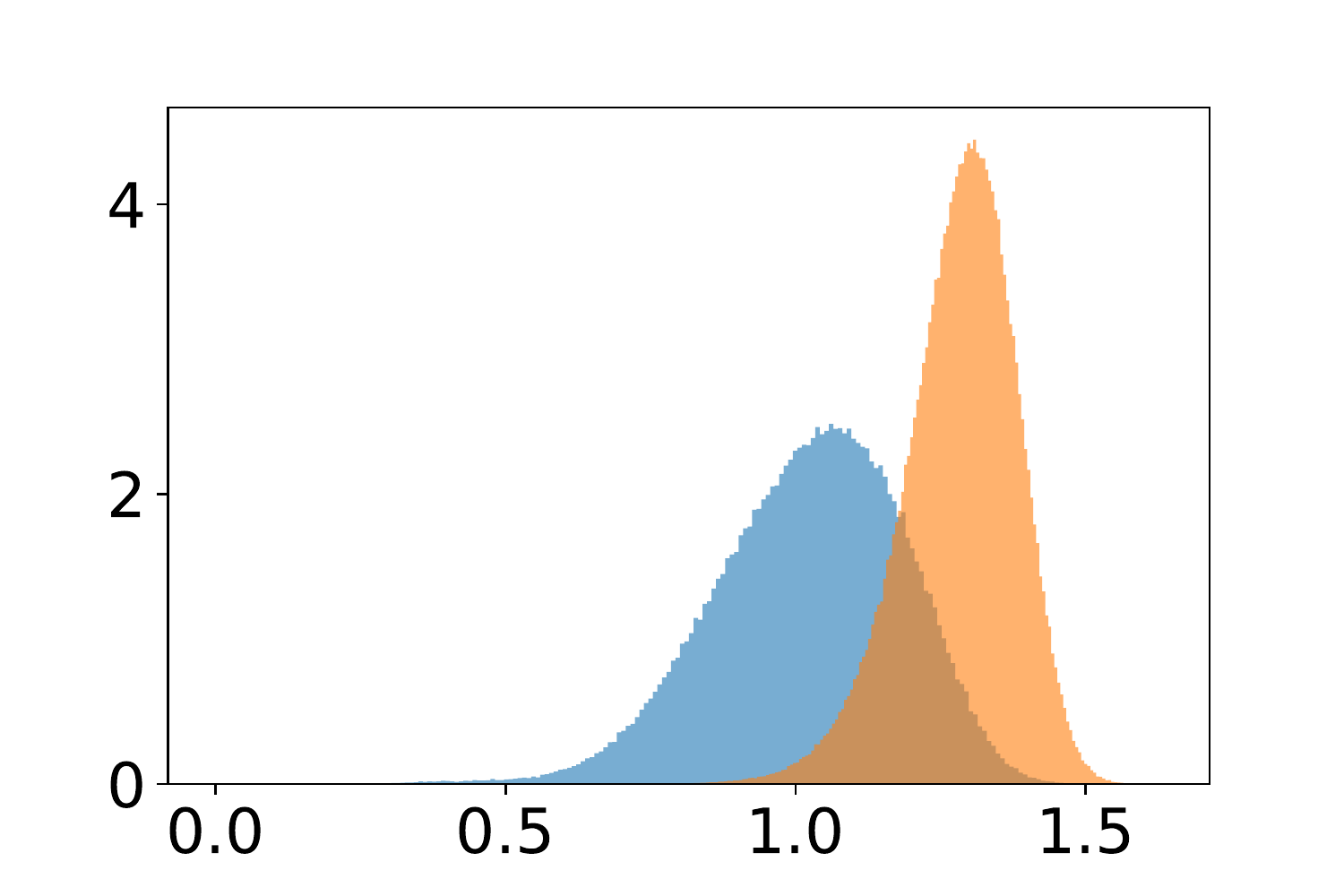} &
\includegraphics[width=0.2\textwidth]{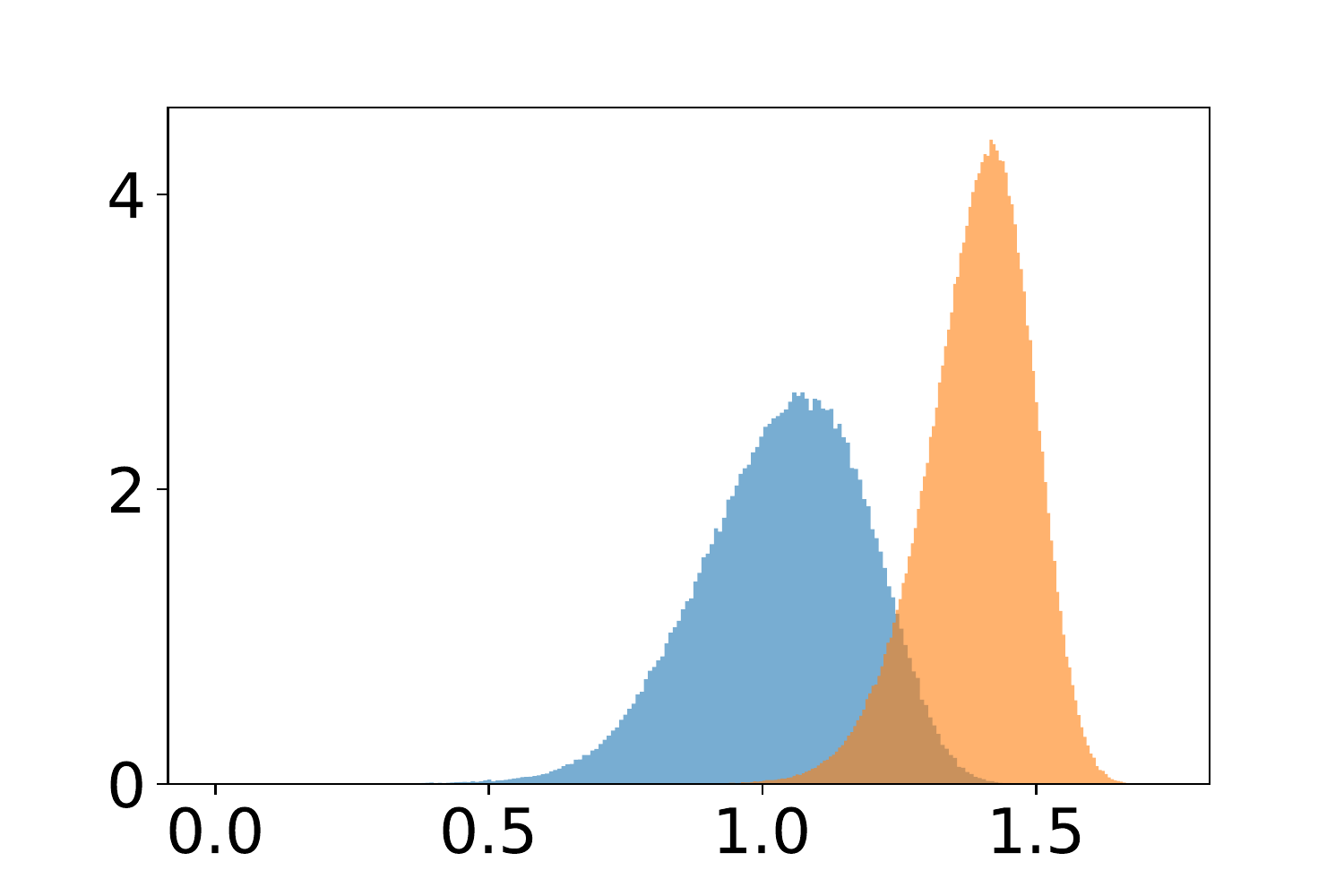}\\
\small (a) &  & (b) &  \\ 
\end{tabular}
\caption{Effect of SFT on feature distributions. (a) \textbf{Left}: Divergences of each class in baseline and SFT. The blue dashed line represents the average divergence of baseline. \textbf{Right}: The standard deviation of divergences in baseline and SFT. 
(b) Histograms of positive(blue) and negative(orange) distance distributions on the Cars196 test set for(from left to right), initial state with pre-trained model, training with ranked list loss, training with ranked list loss together with SFT.
}
\label{fig:divergence_bar}

\end{figure}

\noindent\textbf{Effect on Feature Distributions}
We find that SFT can make feature distributions more similar than the baseline method. This is consistent with our prior that feature distributions should be similar to each other.
In other words, the SFT can make the eigenvalues of variances from different classes to be closer.
In specific, we compare the similarity by comparing the trace of the scattering matrix of each class.
We refer to the trace of the matrix as the divergence. 
The scattering matrix of each class is defined as:
\begin{equation}
{\bf S}_{w,k} = \frac{1}{N_k}\sum_{k=0}^{N_k}\left({\bf x}_k - {\bm \mu}_i\right)^T\left({\bf x}_k - {\bm \mu}_i\right).
\label{eq:scatter_def_individual}
\end{equation}
Fig.~\ref{fig:divergence_bar}(a) shows the divergences of each class, and the standard deviation of the divergences. The class IDs are sorted according to the divergences of the baseline. For clarity, one for every four values is chosen to shown in the histogram.
It is observed that the divergences among classes are more balanced when SFT is applied. 
In general, the divergences below the average(blue dashed line) are increased and those above the average are decreased. The right part of Fig.~\ref{fig:divergence_bar} displays the standard deviations of the divergence values. It is consistent with the conclusion.

Furthermore, the distributions of pair distance are compared. It is shown in
Figure \ref{fig:divergence_bar}(b). It is observed that the overlap between the positive parts and negative parts is reduced when SFT is applied. This indicates that SFT helps the network to learn a more discriminative feature space.

\begin{wraptable}{r}{0.4\textwidth}
\centering
\caption{Effect of SFT on an unbalanced dataset. Head represents classes with rich samples. Tail represents classes in short of training samples. The results are the classification accuracy(\%).}
\label{tab:unbalance}
\resizebox{.4\textwidth}{!}{
\begin{tabular}{cccc}
\hline
             & Baseline & Balanced Train & Unbalanced Train \\ \hline\hline
Head & 95.61    & \bf{95.77}     & 95.49            \\
Tail & 73.91    & 78.35          & \bf{82.32}       \\ \hline
\end{tabular}
}
\end{wraptable}
\noindent\textbf{Effect on Unbalanced Datasets}
The face recognition datasets differ from DML datasets in that they are usually long-tailed. Among then, plenty of classes are in short of samples. These classes are usually called the tail classes. 
Experimentally, we find SFT can improve the performance of tail classes. In specific, we select all classes in MS-Celeb-1M that contains more than 100 samples to construct a mini-dataset. In total, we get 2,445 classes. 
Then, we random choose 1,500 classes to be the head classes and choose 50 samples each for training. For the remaining 945 classes, we treat them as the tail classes and choose 5 samples each for training. All the other samples are left for testing. 
As the training set is much smaller than that of MS-Celeb-1M, we adopt a smaller network for training. In specific, we use a similar CNN architecture for training except that we change the number of residual units to $[1,1,1,1]$. The results are shown in Table~\ref{tab:unbalance}. We can see that the baseline method performs worst in the tail classes. But when SFT is applied, the performance of the tail classes is increased by a large margin. The best performance in tail classes is achieved by the unbalanced train scheme, which outperforms the baseline by 8.4\%, while the performance drop in the head classes is negligible. In summary, the SFT can effectively improve the accuracy of the tail classes.

\begin{wraptable}{r}{0.4\textwidth}
\centering
\caption{Impact of center estimation. 
}
\label{tab:centers}
\resizebox{.4\textwidth}{!}{
\begin{tabular}{ccccc}
\hline
         & baseline & SFT & Random & Pick \\ \hline \hline
R@1 & 74.2     & \textbf{80.2}     & 74.4     & 74.6   \\ \hline
\end{tabular}
}
\end{wraptable}
\noindent\textbf{Impact of Center Estimation}
As the rotation matrix of SFT is estimated based on feature centers, the center estimation is essential. 
To evaluate the importance, we compare the image retrieval performance under three circumstances: (1) ``Random'', skip the center estimation step in line 6 of Algorithm~\ref{algorithm:training_scheme}. (2)``Pick'', randomly pick one sample from the same class as center. (3)SFT, the standard SFT procedure.
The results are shown in Table \ref{tab:centers}.
The performances of SFT are almost the same as the baseline when Random or Pick is adopted. While only when SFT is adopted will the performance be improved by a noticeable number. This illustrate that the accurate estimation of class centers is crucial for feature transform. Moreover, it is noticed that even when the center estimation is not accurate, the feature transform will not harm the training too much. This suggests that training with feature transform is stable.

\begin{wrapfigure}{r}{0.4\textwidth}
\centering
\resizebox{.4\textwidth}{!}{
\begin{tabular}{@{}c@{\hspace{-1mm}}c@{}c@{}}
\includegraphics[height=2.5cm]{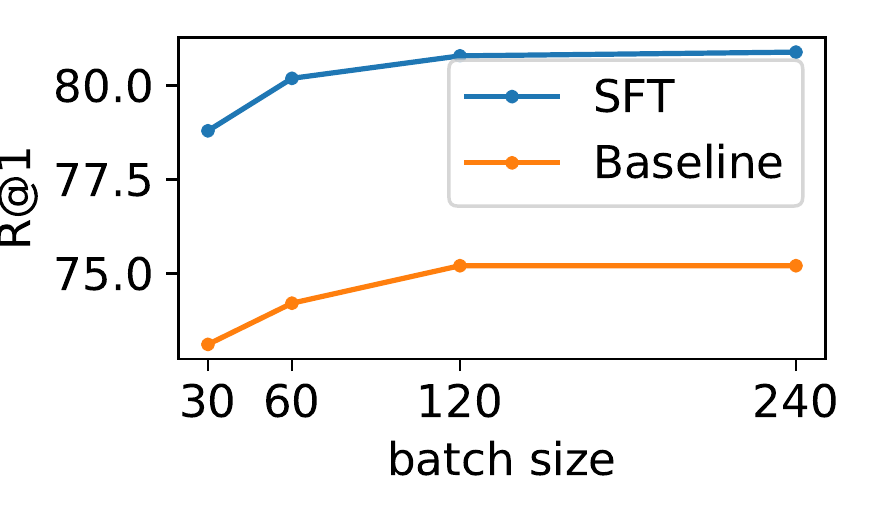} &
\includegraphics[height=2.5cm]{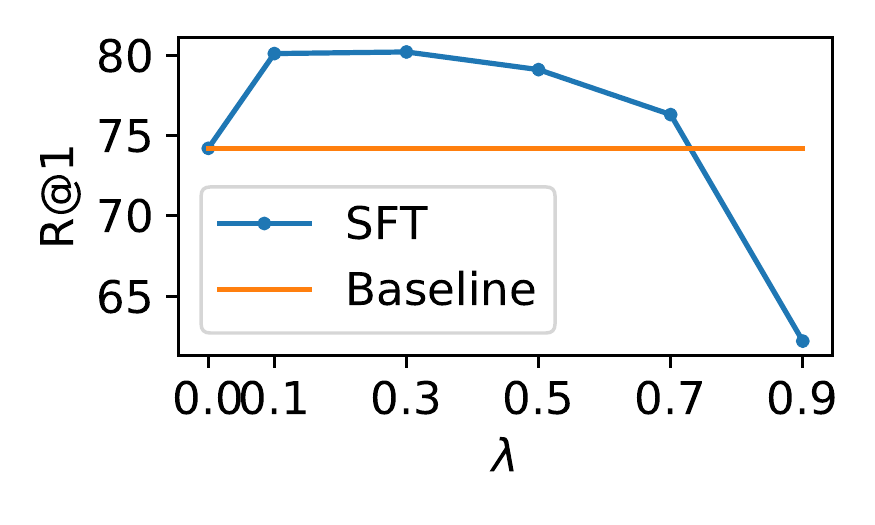} 
\end{tabular}
}
\caption{Performance with different \textbf{Left}: batch size; \textbf{Right} $\lambda$.}
\label{fig:batchsize_lambda}
\end{wrapfigure}

\noindent\textbf{Batch Size.} 
The batch size is usually important in deep metric learning as it determines the number of positive pairs and negative pairs used for constructing target loss. While when implemented with our method, the number of positive pairs and negative pairs are enlarged. We then conduct experiments on different batch sizes to ``fairly'' compare the performance under an equal number of positive and negative pairs. The comparison results are shown in the left part of Fig.~\ref{fig:batchsize_lambda}. It is observed that SFT can beat the baseline with the largest batch size 240 even evaluated under a small batch size 30. This suggests that the improvement when SFT is applied is not due to the increase of batch size.

\noindent\textbf{Effect of $\lambda$. }
We conduct experiments to explore the influence of the weight factor. As shown in the Fig.~\ref{fig:batchsize_lambda}, when increasing the $\lambda$, the performance of the method first increases and then decreases. When the $\lambda$ is too large, the performance drops significantly.
We blame the performance drop to that the gradients from the generated features will dominate the optimization process and infect the optimization of the regular ones. 
In practice, the optimal $\lambda$ is data-dependent. We do not investigate into what is the optimal value of $\lambda$. In most of our experiments, the value is set to 0.2.

\begin{figure}[!tp]
\begin{center}
\begin{tabular}{@{}c@{\hspace{2mm}}c@{}}
\includegraphics[width=.3\textwidth]{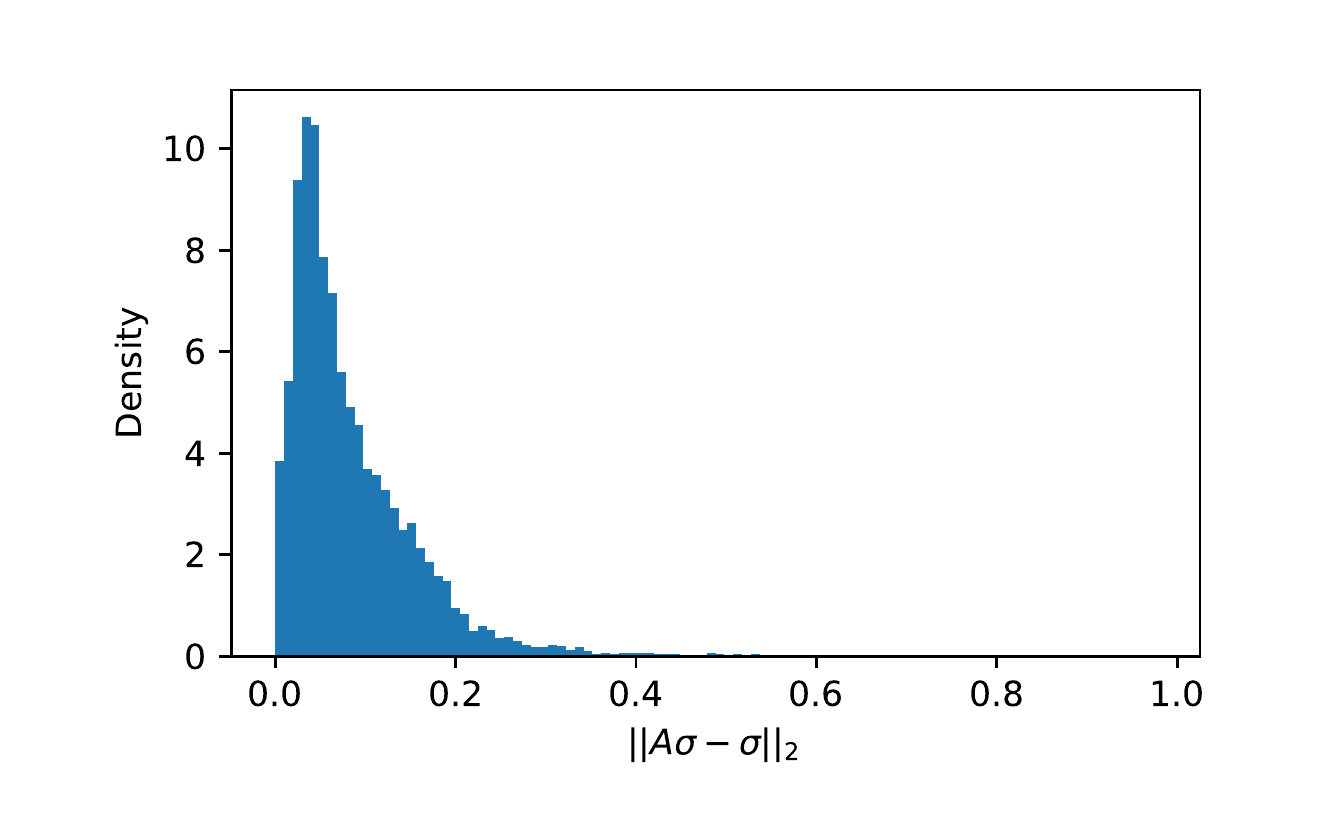}
\includegraphics[width=.3\textwidth]{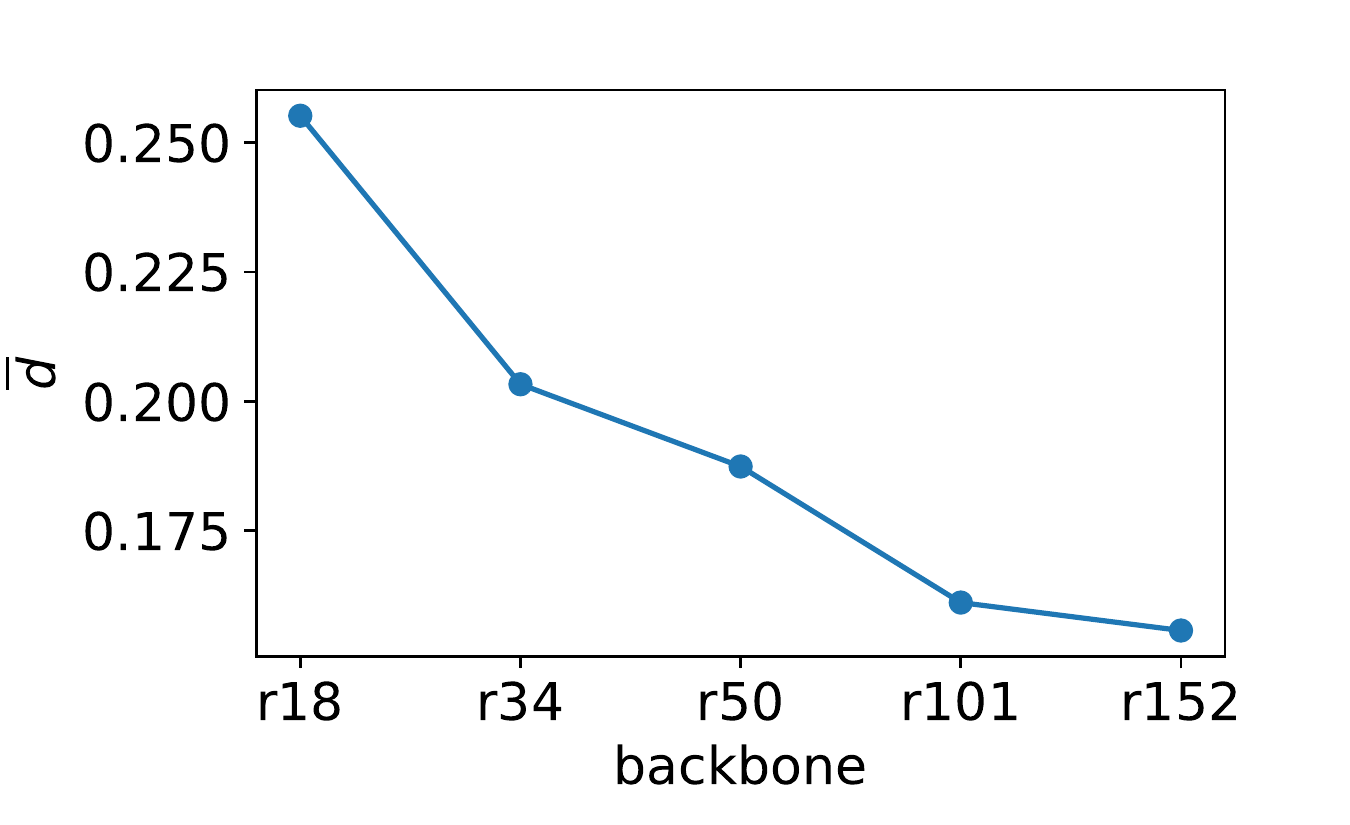}
\includegraphics[width=.3\textwidth]{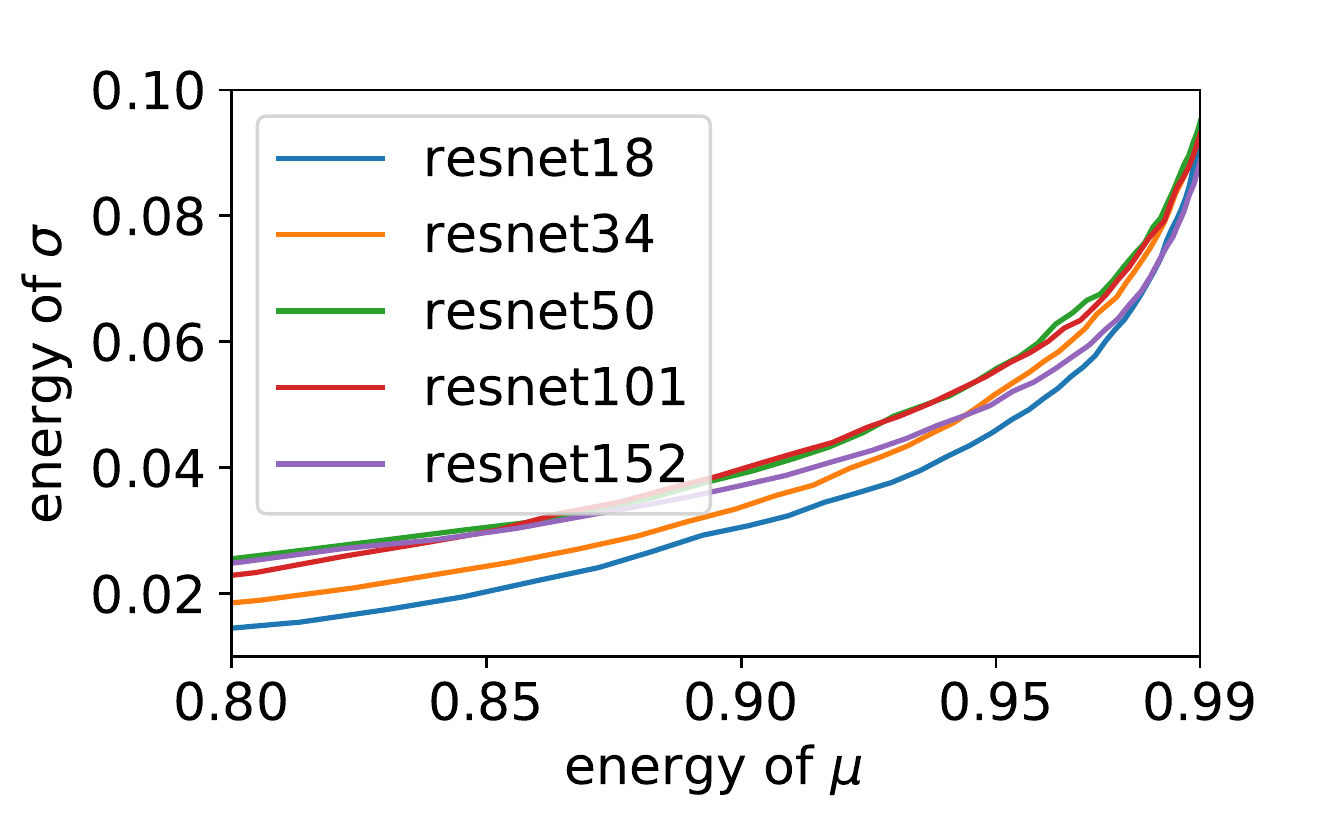}
\end{tabular}
\end{center}
\caption{Experiment on the condition of the degeneration. (a) The distribution of $\|\bf{A}\bm{\sigma}-\bm{\sigma}\|_2$ sampled from features of backbone ResNet50. (b) The mean values of the distributions of $\|\bf{A}\bm{\sigma}-\bm{\sigma}\|_2$ from different backbones. (c) $s_r$ defined in Eq.~\ref{eq:degeneration_ratio} with respect to the energy of $\bm{\mu}$. }
\label{fig:couple_exp}
\end{figure}

\noindent\textbf{Discussion of Degeneration. }
In Sec~\ref{sec:theo}, it is hypothesized that $\bm{\sigma}$ defined in Eq.~\ref{eq:split} is likely to lie in the invariant subspace of $\bf{A}$. To investigate into whether the hypothesis holds, we evaluate the value of $d = \|\bf{A}\bm{\sigma} - \bm{\sigma}\|_2$ on five versions of ResNet. The distribution of $d$ on ResNet50 is shown in Fig.~\ref{fig:couple_exp}(a). It is noticed that a large number of $d$ values are near zero. For these features, the augmented features by SFT and the degenerated form are close. 
For each backbone, $d$ is evaluated $10000$ times and the mean value is reported. The result is shown in Fig.~\ref{fig:couple_exp}(b). As observed, the hypothesis is more likely to hold in deeper networks. This implies that the degeneration of SFT will be more likely to happen when the network gets deeper.

Our experiment also reveals that the learned subspaces for $\bm{\mu}$ and $\bm{\sigma}$ tend to be orthogonal. This is presented in Fig.~\ref{fig:couple_exp}(c). For example, on the backbone of ResNet50, $\bm{\sigma}$ only distribute 10\% energy on the subspace that covers 99\% energy of $\bm{\mu}$. It means that, although the ideal condition in 
Proposition 3
can not be reached, the learned feature space tend to approach it. These experimental results support our analysis that the degeneration of SFT happens for most features. Considering the comparison shows that the SFT will outperform the degenerated form in most scenarios, the side effect of the degenerated form on features that won't degenerate should not be neglected.

\section{Conclusion}
In this paper, we propose Spherical Feature Transform~(SFT) to generate new features from existing ones. The proposed SFT can effectively enrich the intra-class variances of both regular classes and under-represented ones. We have demonstrated the effectiveness of SFT by applying it to several most recent DML frameworks in three popular deep metric learning benchmark datasets and three face recognition benchmark datasets.

\paragraph{\textbf{Acknowledgment. }}
This work was supported in part by the National Key Research and Development Program of China under Grant 2017YFA0700800.
%
%
\bibliographystyle{splncs04}
\bibliography{SFT}
\end{document}